\documentclass[10pt,twocolumn,letterpaper]{article}

\usepackage{iccv}
\usepackage{times}
\usepackage{epsfig}
\usepackage{graphicx}
\usepackage{amsmath}
\usepackage{amssymb}

\usepackage{multirow}
\usepackage{mathrsfs}
\usepackage{enumitem}
\usepackage{makecell}
\usepackage{tabulary}
\usepackage{pifont}
\usepackage{subfigure}

\usepackage[breaklinks=true,bookmarks=false]{hyperref}

\iccvfinalcopy % *** Uncomment this line for the final submission

\def\iccvPaperID{****} % *** Enter the ICCV Paper ID here

\ificcvfinal\fi

\begin{document}

\title{Hierarchy Parsing for Image Captioning}

\author{Ting Yao, Yingwei Pan, Yehao Li, and Tao Mei\\
{\centering JD AI Research, Beijing, China}\\
{\tt\small \{tingyao.ustc, panyw.ustc, yehaoli.sysu\}@gmail.com, tmei@jd.com}
}

\maketitle
\ificcvfinal\thispagestyle{empty}\fi

%%%%%%%%% ABSTRACT

\begin{abstract}
  It is always well believed that parsing an image into constituent visual patterns would be helpful for understanding and representing an image. Nevertheless, there has not been evidence in support of the idea on describing an image with a natural-language utterance. In this paper, we introduce a new design to model a hierarchy from instance level (segmentation), region level (detection) to the whole image to delve into a thorough image understanding for captioning. Specifically, we present a HIerarchy Parsing (HIP) architecture that novelly integrates hierarchical structure into image encoder. Technically, an image decomposes into a set of regions and some of the regions are resolved into finer ones. Each region then regresses to an instance, i.e., foreground of the region. Such process naturally builds a hierarchal tree. A tree-structured Long Short-Term Memory (Tree-LSTM) network is then employed to interpret the hierarchal structure and enhance all the instance-level, region-level and image-level features. Our HIP is appealing in view that it is pluggable to any neural captioning models. Extensive experiments on COCO image captioning dataset demonstrate the superiority of HIP. More remarkably, HIP plus a top-down attention-based LSTM decoder increases CIDEr-D performance from 120.1\% to 127.2\% on COCO Karpathy test split. When further endowing instance-level and region-level features from HIP with semantic relation learnt through Graph Convolutional Networks (GCN), CIDEr-D is boosted up to 130.6\%.
\end{abstract}

\section{Introduction}
Automatic image captioning is the task of generating a natural sentence that correctly reflects the visual content of an image. Practical automatic image description systems have a great potential impact for instance on robotic vision, or helping visually impaired people by transforming visual signals into information that can be communicated via text-to-speech technology. The recent state-of-the-art image captioning methods use to perform ``encoder-decoder" translation \cite{Donahue14,Karpathy:CVPR15,Vinyals14}. A Convolutional Neural Network (CNN) first encodes an image into a feature vector, and a caption is then decoded from this vector, one word at each time step using a Long Short-Term Memory (LSTM) Network. There are variants of approaches arisen from this methodology, for instance, conducting attention on the feature map \cite{Xu:ICML15} or leveraging attributes to augment image features \cite{yao2017boosting}. Regardless of these different versions, today's neural captioning models tend to leverage correlations from training data and produce generic plausible captions, but lack visual understanding on the compositional patterns in images.

\begin{figure}[!tb]
\vspace{-0.25in}
\centering {\includegraphics[width=0.38\textwidth]{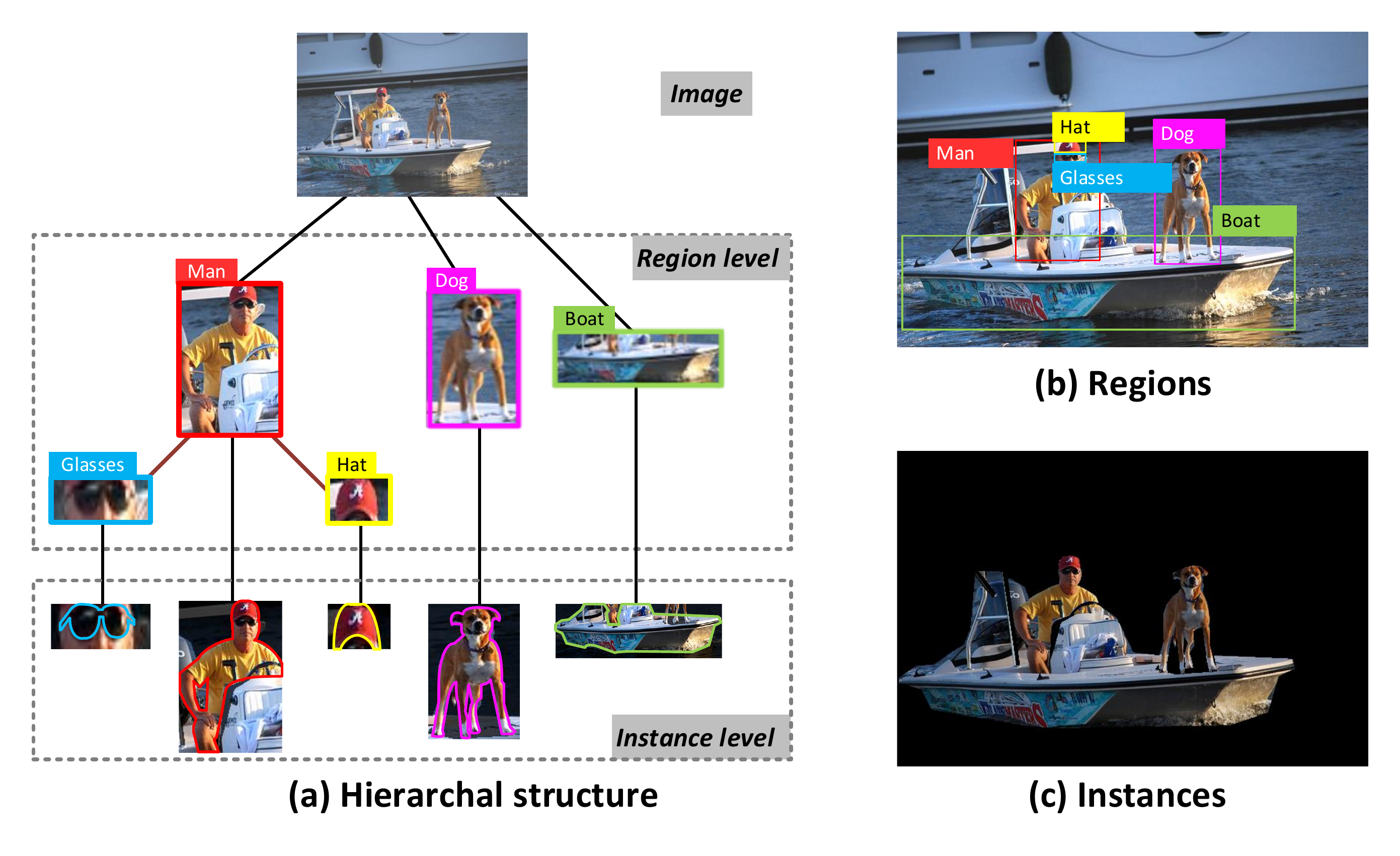}}
\vspace{-0.10in}
\caption{Examples of (a) the hierarchal tree structure in an image, (b) regions and (c) instances in the image.}
\label{fig:fig1}
\vspace{-0.3in}
\end{figure}

We propose to mitigate the problem from the viewpoint of parsing an image into a hierarchical structure of constituent visual patterns to better represent the image. The key idea is to build a top-down hierarchical tree from the root of the whole image to the middle layers of regions and the leaf layer of instances. Each instance in the leaf layer emphasizes the discriminative foreground of a region. Figure \ref{fig:fig1} (a) conceptualizes the typical development of a tree structure on an image. Figure \ref{fig:fig1} (b) and (c) illustrates the regions and foreground/instances in the image, respectively. In this case, we could strengthen visual interpretations of image structure in a bottom-up manner and the learning of all the image-level, region-level and instance-level features does benefit from tree-structured topologies. We expect our design to be a feature refiner or bank, that outputs rich and multi-level representations of the image. Such multi-level representations could be utilized separately or jointly, depending on the particular task. It is also flexible to further integrate the reasoning of semantic relation between regions or instances to further endow region-level or instance-level features with more power.

By consolidating the exploitation of structure hierarchy in an image, we present a new HIerarchy Parsing (HIP) architecture for boosting image encoder in captioning. Specifically, we devise a three-level hierarchy, in which an image is first decomposed into a set of regions and one region is represented either at that level or by further split into finer ones. Each region then corresponds to an instance, that models foreground of the region. A tree-structured LSTM is then executed on the hierarchy from the bottom up to upgrade the features throughout all the levels. After that, a hybrid of features on three levels output by HIP could be easily fed into a general attention-based LSTM decoder to produce the sentence, one word at each time step. Moreover, our HIP, as a feature optimizer, would further augment the features by propagating and aggregating semantic relation. On the hierarchy, we build semantic graph with directed edges on region or instance level, where the vertex represents each region or instance and the edge denotes the relation between each pair of regions or instances. Graph Convolutional Networks (GCN) are exploited to enrich region/instance features with visual relation in the semantic graph. The enhanced features eventually improve image captioning. Please also note that HIP is flexible to generalize to other vision tasks, e.g., recognition.

The main contribution of this work is the parse of hierarchical structure in an image for captioning. The solution also leads to the elegant view of how to build and interpret the hierarchy of an image, and how to nicely integrate such hierarchy into typical neural captioning frameworks, which are problems not yet fully understood in the literature. Our design is viewed as a feature refiner in general and is readily pluggable to any neural captioning models.

\begin{figure*}[!tb]
\vspace{-0.15in}
\centering {\includegraphics[width=0.92\textwidth]{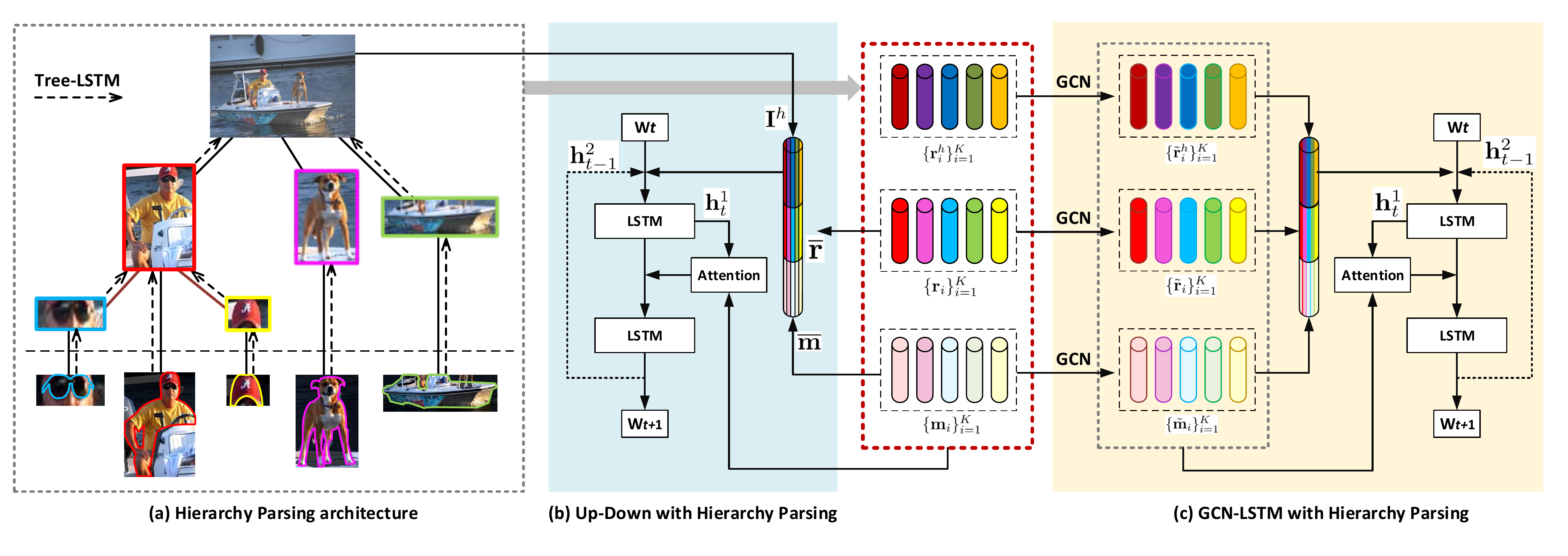}}
\vspace{-0.15in}
\caption{\small An overview of (a) our HIerarchy Parsing (HIP) architecture for integrating hierarchical structure into image encoder, and its applications for image captioning task by plugging HIP in (b) Up-Down \cite{anderson2017bottom} and (c) GCN-LSTM \cite{yao2018exploring}. For HIP, Faster R-CNN and Mask R-CNN are first leveraged to detect and segment the set of object regions and instances, respectively. Next, we construct a three-level hierarchy, where the whole image is first decomposed into a set of regions and one region is represented either at that level or by further split into finer ones. Each region in the middle layers is naturally associated with the corresponding instance in the leaf layer. After that, a Tree-LSTM is executed on the hierarchy from the bottom up with enhanced region/instance features and the outputs are image-level features. A hybrid of features on three levels output by HIP could be easily fed into a general attention-based LSTM decoder in Up-Down for sentence generation. Moreover, it is also flexible to plug our HIP into GCN-LSTM by further enriching the multi-level features with semantic relations in semantic graph via Graph Convolutional Networks (GCN).}
\label{fig:fig2}
\vspace{-0.25in}
\end{figure*}

\section{Related Work}\label{sec:RW}

\textbf{Image Captioning.} The recent works for image captioning \cite{anderson2017bottom,Donahue14,Vinyals14,Xu:ICML15,yao2017boosting,You:CVPR16} are mainly sequence learning based methods which utilize CNN plus RNN to generate sentences word-by-word, enabling the flexible modeling of syntactical structure within sentence. Specifically, \cite{Vinyals14} is one of the early attempts to cast the sentence generation task as a sequence learning problem and leverages LSTM to model the dependency among words for sentence generation conditioning on the input image. \cite{Xu:ICML15} further extends \cite{Vinyals14} by integrating soft and hard attention mechanism into LSTM-based decoder, which learns to focus on image regions to facilitate the generation of the corresponding word at each decoding stage. \cite{Wu:CVPR16,yao2017boosting,You:CVPR16} demonstrate the effectiveness of semantic attributes in image captioning, where the attributes are taken as additional inputs of CNN plus RNN to emphasize them in output sentence. Later on, \cite{rennie2017self} develops a self-critical sequence training strategy to amend the discrepancy between training and inference for sequence modeling and thus boost image captioning. Furthermore, instead of measuring attention over a pre-defined uniform grid of image regions as in \cite{Xu:ICML15}, \cite{anderson2017bottom} especially devises the bottom-up mechanism to enable the measurement of attention at object level, and the top-down mechanism to associate the salient image regions and the output words for sentence generation. Most recently, \cite{yao2018exploring} models the relations between objects in the context of image captioning, which will be further incorporated into the top-down attention model \cite{anderson2017bottom} to enhance captions. In addition, image captioning could be extended to novel object captioning \cite{li2019novel,yao2017deep} which leverages unpaired image/text data to describe novel objects or image paragraph generation \cite{wang2019paragraph} which produces a coherent paragraph to depict an image.

In our work, we exploit the hierarchal structure in images from instance level, region level, to the whole image, to facilitate a thorough image understanding for captioning. To do this, we design a novel hierarchy parsing architecture to integrate hierarchical structure into image encoder, which is pluggable to any neural captioning models.

\textbf{Structured Scene Parsing.} The task of structured scene parsing goes beyond the general recognition of scene type (classification) or localization of objects in a scene (semantic labeling or segmentation) and considers a deeper and structured understanding on scene. An early pioneering work \cite{tu2005image} devises a Bayesian framework for parsing images into the constituent visual patterns over a hierarchical parsing graph. Later on, Han \emph{et al.} leverage an attributed grammar model to hierarchically parse the man-made indoor scene \cite{han2009bottom}. A connected segmentation tree is proposed in \cite{ahuja2008connected} to capture canonical characteristics of the object in terms of photometric and geometric properties, and containment and neighbor relationships between its constituent image regions. \cite{zhu2012recursive} designs a hierarchical image model for image parsing, which represents image structures with different levels of contextual information. In \cite{liu2015understanding}, a hierarchical shape parsing strategy is proposed to partition and organize image components into a hierarchical structure in the scale space. Sharma \emph{et al.} devise the recursive context propagation network \cite{sharma2014recursive} for semantic scene labeling by recursively aggregating contextual information from local neighborhoods up to the entire image and then disseminating the aggregated information back to individual local features over a binary parse tree.

The hierarchy parsing architecture in our method is also a type of structured scene parsing for images. Unlike the aforementioned methods that are developed for image parsing or semantic scene labeling, our hierarchy parsing architecture acts as an image encoder to interpret the hierarchal structure in images and is applicable to image captioning task. As such, all the instance-level, region-level and image-level features are enhanced with tree-structured topologies, which will be injected into captioning model to further boost sentence generation.

\section{Our Approach}
In this paper, we devise a HIerarchy Parsing (HIP) architecture to integrate hierarchical structure into image encoder, pursuing a thorough image understanding to facilitate image captioning. HIP firstly constructs a three-level hierarchy from the root of the whole image to the middle layers of regions and the leaf layer of instances, leading to a deep and structured modeling of image. Tree-LSTM is then leveraged to contextually enhance features at instance level, region level, and image level. In this sense, HIP acts as a feature refiner, that outputs rich and multi-level representations of an image. Hence HIP is pluggable to any neural captioning models, including the general attention-based LSTM decoder or a specific relation-augmented decoder. An overview of our HIP architecture and its applications in two different captioning models is depicted in Figure \ref{fig:fig2}.

\subsection{Overview}
\textbf{Notation.} The target of image captioning task is to describe the given image ${I}$ with a textual sentence $\mathcal {S}$. Note that the textual sentence $\mathcal{S} = \{w_1, w_2, ..., w_{N_s}\}$ is a word sequence containing $N_s$ words. Each word in sentence is represented as a $D_s$-dimensional textual feature, e.g., ${\bf{w}}_t\in {{\mathbb{R}}^{D_s}}$, which denotes the feature of $t$-th word in sentence $\mathcal{S}$. Since our ultimate hierarchy consists of compositional patterns at instance level, region level and image level, we firstly leverage object detection method (Faster R-CNN \cite{ren2015faster}) and instance segmentation approach (Mask R-CNN \cite{he2017mask}) to produce the basic elements (i.e., regions and instances) of the hierarchy. The set of regions and instances in image ${I}$ is denoted as $\mathcal{R}=\{{r_i}\}^{K}_{i=1}$ and $\mathcal{M}=\{{m_i}\}^{K}_{i=1}$ respectively, which corresponds to the regions and foregrounds/instances of detected $K$ objects. Each image region and its instance are denoted as the $D_r$-dimensional features ${{\bf{r}}_i}\in {{\mathbb{R}}^{D_r}}$ and ${{\bf{m}}_i}\in {{\mathbb{R}}^{D_r}}$. Accordingly, the image ${I}$ is parsed into a hierarchal tree $\mathcal{T}=({I}, \mathcal{R},\mathcal{M},\mathcal{E}_{tree})$ consisting of layers at three levels: the root layer corresponds to entire image ${I}$, the middle layers of regions $\mathcal{R}$, and the leaf layer of instances $\mathcal{M}$. $\mathcal{E}_{tree}$ denotes the connections. More details about how we represent regions \& instances and construct hierarchal tree will be elaborated in Section~\ref{sec:SG}.

\textbf{Problem Formulation.} The key idea behind our formulation is to frame the hierarchical structure modeling of image in the context of image captioning task. To start, given the sets of regions and instances decomposed from input image, we holistically characterize an image with a three-level hierarchy. Derived from the idea of tree-structured LSTM \cite{tai2015improved}, we further leverage a Tree-LSTM module to contextually refine the representation of each instance/region in a bottom-up manner along the hierarchical tree and finally acquire the image-level feature. As such, by taking the hierarchy parsing architecture (i.e., the construction of hierarchical tree and the feature enhancement via a Tree-LSTM) as a process of image encoding, the output multi-level representations are endowed with more power. After that, the contextually refined features on three levels from the hierarchy parsing architecture are fed into a general attention-based LSTM decoder \cite{anderson2017bottom} to facilitate sentence generation. Therefore, the image captioning problem here is generally formulated as the minimization of energy loss function:
\begin{equation}\label{Eq:Eq1}\small
E(\mathcal{T}, {\mathcal {S}}) = -\log {\Pr{({\mathcal {S}}|\mathcal{T})}},
\end{equation}
which is the negative log probability of the correct sentence ${\mathcal {S}}$ given the hierarchal tree $\mathcal{T}$. Moreover, since the hierarchy parsing architecture is designed to be a feature refiner or bank, we could further augment the output enhanced features by propagating and aggregating semantic relations. That is, our hierarchy parsing architecture can be applied in the relation-based captioning model \cite{yao2018exploring} which endows the region-level or instance-level features with relations.

\subsection{Hierarchy Parsing in Images} \label{sec:SG}

Next we describe the details of our HIerarchy Parsing (HIP) architecture, which strengthens all the instance-level, region-level and image-level features with tree-structured topologies as a feature optimizer. In particular, we begin this section by presenting the extraction of regions and instances within images. Then, we provide how to construct the three-level hierarchy by associating all regions and instances in an image. Finally, an image encoder equipped with Tree-LSTM for interpreting the hierarchal structure and enhancing multi-level features is presented.

\begin{figure}[!tb]
\vspace{-0.12in}
\centering {\includegraphics[width=0.45\textwidth]{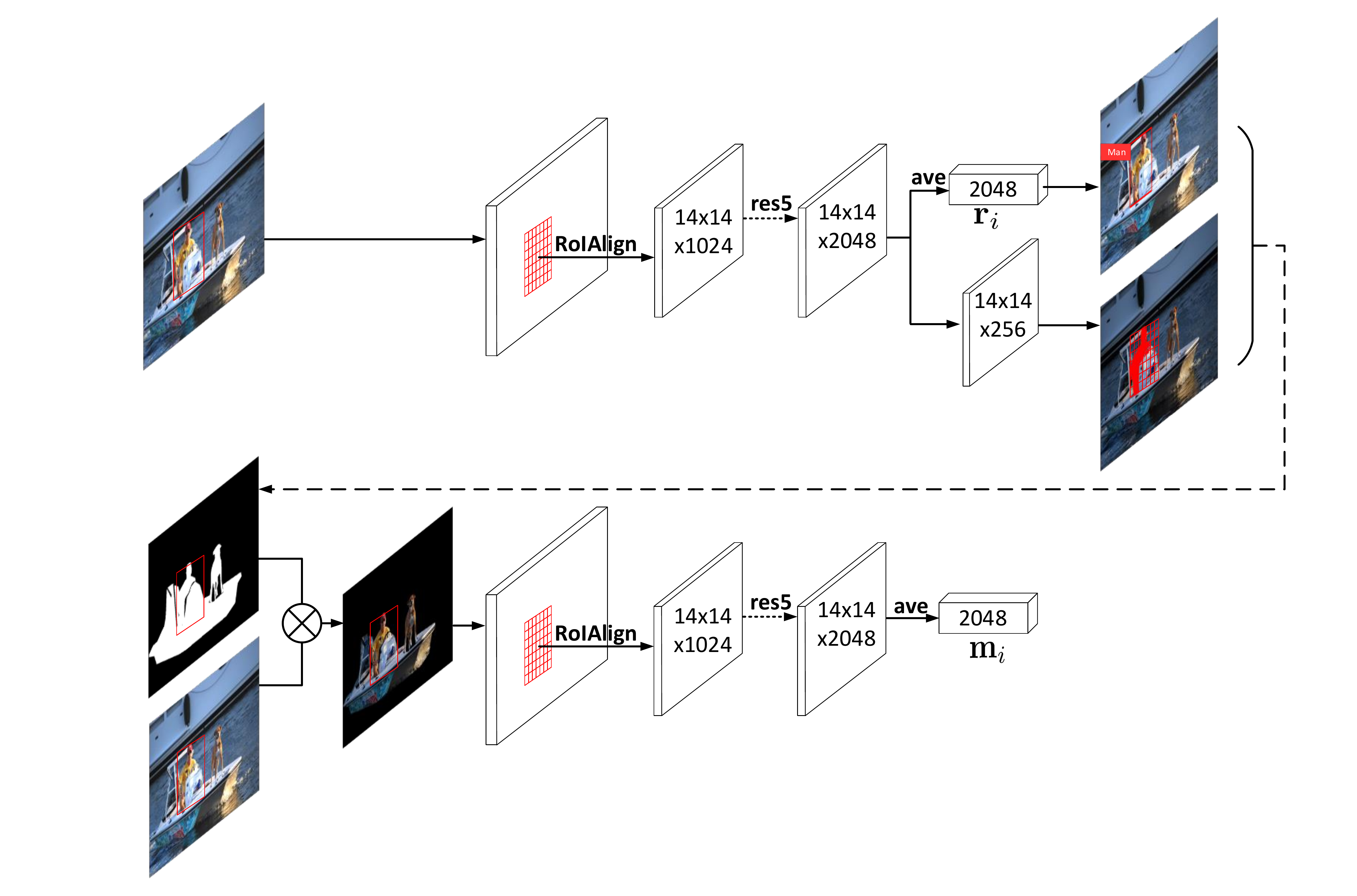}}
\vspace{-0.05in}
\caption{\small Feature extraction of regions and instances. Mask R-CNN augments the pre-trained Faster R-CNN with an additional mask branch and is adopted to extract region feature ${{\bf{r}}_i}$ and predict the instance-level foreground mask of each region. Next, a blend of each region and its binary via element-wise multiplication is fed into another Faster R-CNN to produce instance feature ${{\bf{m}}_i}$.}
\label{fig:figfea}
\vspace{-0.25in}
\end{figure}

\textbf{Regions and Instances of Image.}
Given an input image, we firstly apply Faster R-CNN trained on Visual Genome \cite{krishna2017visual} to detect the image regions of objects. Note that only the top $K=36$ regions with highest confidences $\mathcal{R}=\{{r_i}\}^{K}_{i=1}$ are selected to represent the image. We represent each region as the 2,048-dimensional output (${{\bf{r}}_i}$) of pool5 layer after RoI pooling from the feature map of Faster R-CNN (backbone: ResNet-101 \cite{He:CVPR16}). In addition, to emphasize the discriminative knowledge of objects implicit in each image region, we separate the foreground \& background of each region and take the foreground of region as the associated instance. Specifically, Mask R-CNN augments the pre-trained Faster R-CNN with an additional mask branch and predicts the instance-level foreground mask of each region. As such, the foreground/instance is obtained by blending each region and its binary mask via element-wise multiplication, leading to the set of instances $\mathcal{M}=\{{m_i}\}^{K}_{i=1}$. Next, we train another Faster R-CNN over the foreground images and the 2,048-dimensional output (${{\bf{m}}_i}$) of this Faster R-CNN is taken as the representation of each instance $m_i$. Note that the inputs to the two Faster R-CNN models are different (one is original image and the other is foreground image) and the two models don't share any parameters. Figure \ref{fig:figfea} details the pipeline for the feature extraction of regions and instances.

\textbf{Hierarchy Structure of an Image.}
Recent advances on visual relationship \cite{yao2018exploring} have demonstrated that modeling the structure in an image (e.g., a semantic graph built based on relations between regions) does enhance image captioning. Our work takes a step forward and constructs a hierarchical structure of constituent visual patterns, i.e., hierarchal tree, to fully exploit a hierarchy from instance level, region level to the whole image and learn the connections across each level for image captioning. Specifically, the hierarchal tree $\mathcal{T}=({I}, \mathcal{R},\mathcal{M},\mathcal{E}_{tree})$ organizes all the regions and instances of input image ${I}$ in a top-down three-level hierarchy, including the root layer, the middle layers of regions, and the leaf layer of instances.

Concretely, only one root node is initially established in the upper root layer, which denotes the entire image ${I}$. Such image-level root node is decomposed into a set of regions, which in turn resolve into a number of finer regions, resulting in multiple middle layers of regions. Here the existence of dependency between pairs of root node and regions is assigned depending on their Intersection over Union (IoU). More precisely, given the $K$ image regions $\mathcal{R}$, we firstly rank all the regions in order of descending area of region and then integrate each region into the hierarchal tree in turn. For each region $r_i$, we measure IoU between $r_i$ and each object node in current hierarchal tree. If  the maximum IoU is larger than a threshold $\epsilon$, $r_i$ is incorporated into the hierarchal tree as a child of the existing region node with maximum IoU, which indicates that $r_i$ can be treated as the finer region within the region of its parent. Otherwise, $r_i$ is directly taken as the child node of the image-level root node. Once the construction of middle layers of regions completes, we attach each instance to the corresponding region node as a child in the bottom leaf layer.

\begin{figure}[!tb]
\vspace{-0.15in}
\centering {\includegraphics[width=0.36\textwidth]{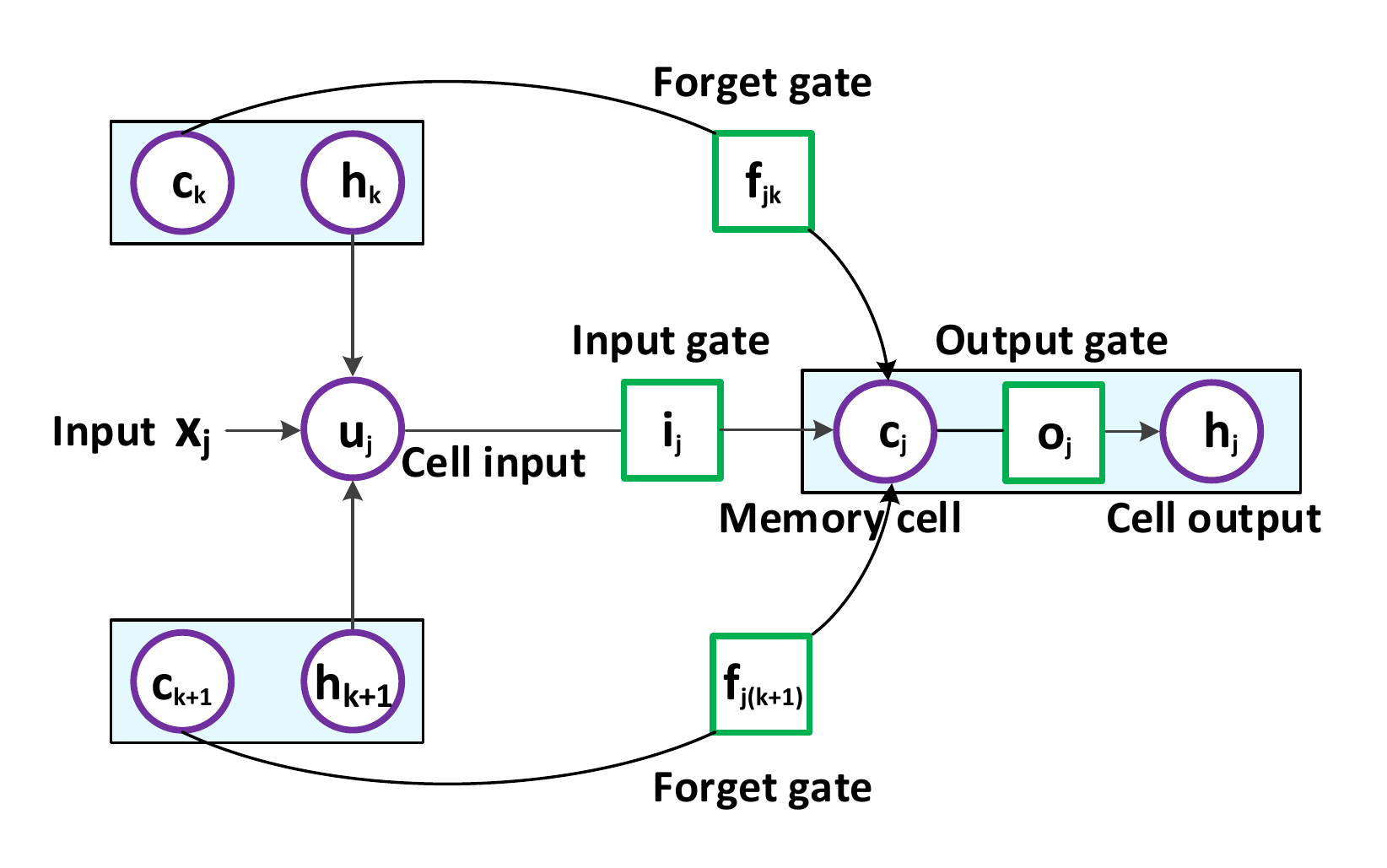}}
\vspace{-0.15in}
\caption{\small A diagram of a memory cell $c_j$ with two children (subscripts $k$ and $k+1$) in Tree-LSTM. We omit the dependencies of four gates for compactness.}
\label{fig:fig3}
\vspace{-0.25in}
\end{figure}

\textbf{Image Encoder with Tree-LSTM.}
One natural way to model the contextual relationship across samples in a set/sequence is to adopt LSTM based models as in \cite{Bahdanau14}. However, such kind of chain-structured LSTM is typically order-insensitive and thus insufficient to fully capture the differences in order or dependency structure. Taking the inspiration from the success of Tree-LSTM \cite{tai2015improved} for modeling tree-structured topologies in several NLP tasks, we leverage Tree-LSTM in image encoder to facilitate contextual information mining within hierarchy and thus enrich image-level features with holistic hierarchical structure.

A diagram of the Tree-LSTM unit is illustrated in Figure \ref{fig:fig3}. Similar to the standard LSTM, Tree-LSTM unit consists of a memory cell ${{\bf{c}}_j}$ indexed by $j$, hidden state ${{\bf{h}}_j}$, input gate ${{\bf{i}}_j}$, and output gate ${{\bf{o}}_j}$. Unlike LSTM updates memory cell depending on only previous hidden state, the updating of Tree-LSTM unit relies on the multiple hidden states of its children. Moreover, Tree-LSTM unit includes forget gate ${{\bf{f}}_{jk}}$ for each child (indexed by $k$). In particular, the vector formulas for a Tree-LSTM unit forward pass are given below. For the node indexed by $j$ in a tree, ${\bf{x}}_j$ and ${\bf{h}}_j$ denote the input and output vector, respectively. The set of children of this node is denoted as $C(j)$. $\bf{W}$ are input weights matrices, $\bf{U}$ are recurrent weight matrices and $\bf{b}$ are biases. Sigmoid $\sigma$ and hyperbolic tangent $\phi$ are element-wise non-linear activation functions. $\odot$ represents the dot product of two vectors. Hence the Tree-LSTM unit updates~are:
\vspace{-0.05in}
\begin{equation}\label{Eq:Eqtlstm}\small
\begin{array}{l}
{{\bf{\widetilde{h}}}_j} ~= \sum\limits_{k \in C(j)} {{\bf{h}}_k} \\
{{\bf{u}}_j} ~= \phi ({{\bf{W}}_u}{{\bf{x}}_j} + {{\bf{U}}_u}{{\bf{\widetilde{h}}}_j} + {{\bf{b}}_u}) ~~~~~~~~~~~cell~input\\
{{\bf{i}}_j} ~~= \sigma ({{\bf{W}}_i}{{\bf{x}}_j} + {{\bf{U}}_i}{{\bf{\widetilde{h}}}_j} + {{\bf{b}}_i}) ~~~~~~~~~~~~input~gate \\
{{\bf{f}}_{jk}} = \sigma ({{\bf{W}}_f}{{\bf{x}}_j} + {{\bf{U}}_f}{{\bf{h}}_k} + {{\bf{b}}_f}) ~~~~~~~forget~gate \\
{{\bf{o}}_j} ~= \sigma ({{\bf{W}}_o}{{\bf{x}}_j} + {{\bf{U}}_o}{{\bf{\widetilde{h}}}_j} + {{\bf{b}}_o}) ~~~~~~~~output~gate \\
{{\bf{c}}_j} ~= {\bf{u}}_j \odot {\bf{i}}_j + \sum\limits_{k \in C(j)} {\bf{c}}_k \odot {\bf{f}}_{jk} ~~~~~~~~~~cell~state \\
{{\bf{h}}_j} ~= \phi ({\bf{c}}_j) \odot {\bf{o}}_j ~~~~~~~~~~~~~~~~~~~~~~~~~~~~~~~cell~output
\end{array}.
\end{equation}
Specifically, for our hierarchal tree $\mathcal{T}$, we take the original extracted region/instance features ($\{{\bf{r}}_i\}^{K}_{i=1}$ and $\{{\bf{m}}_i\}^{K}_{i=1}$) as the input vectors of region nodes in the middle layers and instance nodes in the leaf layer. The input vector of image-level root node is set as the linear fusion of image-level mean-pooled features of regions ({\small $\overline {\bf{r}} = \frac{1}{K}\sum\nolimits_{i = 1}^K {{\bf{r}}_i}$}) and instances ({\small $\overline {\bf{m}} = \frac{1}{K}\sum\nolimits_{i = 1}^K {{\bf{m}}_i}$}): ${\bf{I}}={{\bf{W}}_r}\overline {\bf{r}}+{{\bf{W}}_m}\overline {\bf{m}}$. Accordingly, by operating Tree-LSTM over hierarchal tree in a bottom-up manner, the region-level features of each region node are further strengthened with the contextual information mined from its instance and even finer regions, which are denoted as $\{{\bf{r}}^h_i\}^{K}_{i=1}$. In addition, the outputs of root node in hierarchal tree are treated as the image-level features ${\bf{I}}^h$, which are endowed with inherent hierarchal structure from instance level, region level to the whole image.

\subsection{Image Captioning with Hierarchy Parsing}

Since we design our HIP architecture to be a feature refiner or bank that outputs rich and multi-level representations of the image, it is feasible to plug HIP into any neural captioning models. We next discuss how to integrate hierarchy parsing into a general attention-based LSTM decoder in Up-Down \cite{anderson2017bottom} or a specific relation-augmented decoder in GCN-LSTM \cite{yao2018exploring}. Please also note that our HIP is flexible to generalize to other vision tasks, e.g., recognition.

\textbf{Up-Down with Hierarchy Parsing.}
Given a hybrid of features on three levels output by HIP (i.e., the image-level features (${\bf{I}}^h$, $\overline {\bf{r}}$, $\overline {\bf{m}}$) and region-level/instance-level features ($\{{\bf{r}}^h_i\}^{K}_{i=1}$, $\{{\bf{r}}_i\}^{K}_{i=1}$, $\{{\bf{m}}_i\}^{K}_{i=1}$), we directly feed them into a general attention-based decoder with two-layer LSTM in Up-Down, as depicted in Figure \ref{fig:fig2} (b). Specifically, at each time step $t$, the input of the first-layer LSTM unit is set as the concatenation of the input word $w_t$, the previous output of the second-layer LSTM unit ${\bf{h}}^2_{t-1}$ and all image-level features (${\bf{I}}^h$, $\overline {\bf{r}}$, $\overline {\bf{m}}$). Such design can collect the maximum contextual information for the first-layer LSTM to model dependency among words. After that, we represent each image region by concatenating all region-level and instance-level features belonging to it, denoted as ${\bf{v}}_i=\left[{\bf{r}}^h_i, {\bf{r}}_i, {\bf{m}}_i \right]$. Next, a normalized attention distribution $\lambda_t \in\mathbb R^{K}$ over all regions $\{{\bf{v}}_i\}^{K}_{i=1}$ is calculated conditioning on the output ${\bf{h}}_t^1$ of the first-layer LSTM unit, resulting in the attended image feature {\small $\hat {\bf{v}}_t  = \sum\nolimits_{i = 1}^K {\lambda_{t,i}{\bf{v}}_i}$}. Note that $\lambda_{t,i}$ is the $i$-th element in $\lambda_{t}$ which represents the attention probability of $i$-th region. Therefore, we feed the concatenation of the attended image feature $\hat {\bf{v}}_t$ and ${\bf{h}}_t^1$ into the second-layer LSTM unit, aiming to trigger the generation of next word $w_{t+1}$.

\textbf{GCN-LSTM with Hierarchy Parsing.}
When applying the hierarchy parsing into GCN-LSTM \cite{yao2018exploring}, the instance-level and region-level features from HIP are further enhanced with visual relation learnt through GCN and thus improve the captions, as shown in Figure \ref{fig:fig2} (c). In particular, we firstly build semantic graph with directed edges on region or instance level of the hierarchy. GCN is then leveraged to enrich the region-level/instance-level features ($\{{\bf{r}}^h_i\}^{K}_{i=1}$, $\{{\bf{r}}_i\}^{K}_{i=1}$, $\{{\bf{m}}_i\}^{K}_{i=1}$) with visual relations in the semantic graph. All of the enhanced region-level/instance-level features ($\{\tilde{\bf{r}}^h_i\}^{K}_{i=1}$, $\{\tilde{\bf{r}}_i\}^{K}_{i=1}$, $\{\tilde{\bf{m}}_i\}^{K}_{i=1}$) from GCN are further fed into a two-layer LSTM for sentence generation.

\textbf{Extension to Recognition Task.}
The image-level features from our HIP can be further utilized to other vision tasks, e.g., recognition. The spirit behind follows the philosophy that the hierarchy parsing integrates hierarchical structure of image into encoder, making the learnt image-level features more representative and discriminative.

\section{Experiments}
We empirically verify the merit of our HIP by conducting experiments on COCO \cite{Lin:ECCV14} for image captioning task.

\begin{table*}[t]\small
    \centering
    \setlength\tabcolsep{2.5pt}
    \caption{\small Performance (\%) of our HIP and other methods on COCO Karpathy test split.}
    \vspace{-0.0in}
    \begin{tabular}{l | c c c c c | c c c c c}
        \Xhline{2\arrayrulewidth}
		  & \multicolumn{5}{c|}{\textbf{Cross-Entropy Loss}} & \multicolumn{5}{c}{\textbf{CIDEr-D Score Optimization}} \\
		                      & BLEU@$4$    & METEOR       & ROUGE-L      & CIDEr-D       & SPICE         & BLEU@$4$    & METEOR       & ROUGE-L      & CIDEr-D       & SPICE    \\
	      \hline \hline
      LSTM \cite{Vinyals14}                 & ~29.6~ & ~25.2~ & ~52.6~ &  ~94.0~ & ~-~  & ~31.9~ & ~25.5~ & ~54.3~ & ~106.3~ & -    \\
      SCST \cite{rennie2017self}            &  30.0  &  25.9  &  53.4  &   99.4  & -    &  34.2  &  26.7  &  55.7  &  114.0  & -    \\
      ADP-ATT \cite{Xiong2016MetaMind}      &  33.2  &  26.6  &  -     &  108.5  & -    &  -     &  -     &   -    &  -      & -    \\
      LSTM-A \cite{yao2017boosting}         &  35.2  &  26.9  &  55.8  &  108.8  & 20.0 &  35.5  &  27.3  &  56.8  &  118.3  & 20.8 \\
      RFNet    \cite{jiang2018recurrent}    &  37.0  &  27.9  &  57.3  &  116.3  & 20.8 &  37.9  &  28.3  &  58.3  &  125.7  & 21.7 \\\hline
       Up-Down \cite{anderson2017bottom}    &  36.2  &  27.0  &  56.4  &  113.5  & 20.3 &  36.3  &  27.7  &  56.9  &  120.1  & 21.4 \\
      Up-Down+HIP                           &  37.0  &  28.1  &  57.1  &  116.6  & 21.2 &  38.2  &  28.4  &  58.3  &  127.2  & 21.9 \\\hline
      GCN-LSTM   \cite{yao2018exploring}    &  37.1  &  28.1  &  57.2  &  117.1  & 21.1 &  38.3  &  28.6  &  58.5  &  128.7  & 22.1 \\
      GCN-LSTM+HIP                          &  \textbf{38.0}  &  \textbf{28.6}  &  \textbf{57.8}  &  \textbf{120.3}  & \textbf{21.4} &  \textbf{39.1}  &  \textbf{28.9}  &  \textbf{59.2}  & \textbf{130.6}   & \textbf{22.3} \\
		\Xhline{2\arrayrulewidth}
    \end{tabular}
	\vspace{-0.22in}
    \label{tab:COCO}
\end{table*}

\subsection{Datasets and Settings}
\textbf{COCO} is a standard benchmark in the field of image captioning. The dataset contains 123,287 images (82,783 for training and 40,504 for validation) and each image is annotated with 5 descriptions. Given the fact that the human-annotated descriptions of the official testing set are not provided, we utilize Karpathy split (113,287 for training, 5,000 for validation and 5,000 for testing) as in \cite{anderson2017bottom}. Following \cite{Karpathy:CVPR15}, all the training sentences are converted to lower case and we omit rare words which occur less than 5 times. As such, the final vocabulary includes 10,201 unique words.

\textbf{Visual Genome} is adopted to train Faster R-CNN for object detection. Here we follow the setting in \cite{anderson2017bottom,yao2018exploring} and take 98,077 images for training, 5,000 for validation, and 5,000 for testing. As in \cite{anderson2017bottom}, 1,600 objects and 400 attributes are selected from Visual Genome for training Faster R-CNN with two branches for predicting object and attribute classes.

\textbf{COCO-detect} is a popular benchmark for instance segmentation, containing the same images with COCO from 80 object categories. All object instances are annotated with a detailed segmentation mask. Here we utilize the partially supervised training paradigm \cite{hu2017learning} to train Mask R-CNN, enabling instance segmentation over the entire 1,600 objects. In particular, the detection branch in Mask R-CNN is initialized with the weights of learnt Faster R-CNN from Visual Genome. Next, the mask branch and weight transfer function in Mask R-CNN are further trained on COCO-detect. Note that we adopt the same split of COCO for training Mask R-CNN on COCO-detect.

\textbf{Implementation Details.}
We represent each word as ``one-hot" vector. The threshold $\epsilon$ for constructing hierarchy is set as $0.1$. The hidden layer size in Tree-LSTM and LSTM-based decoder is set as 500 and 1,000, respectively. The captioning models with our HIP are mainly implemented with PyTorch, optimized with Adam \cite{kingma2014adam}. For the training with cross-entropy loss, we set the learning rate as $5\times10^{-4}$ and the mini-batch size as 50. The maximum iteration is set as 30 epoches. For the training with self-critical training strategy, we follow \cite{rennie2017self} and select the model which is trained with cross-entropy loss and achieves best CIDEr-D score on validation set, as initialization. Next the captioning model is further optimized with CIDEr-D reward. Here the learning rate is set as $5\times10^{-5}$ and the maximum iteration is 30 epoches. At inference, beam search strategy is adopted and we set the beam size as 3. Five popular metrics, i.e., BLEU@$N$ \cite{Papineni:ACL02}, METEOR \cite{Banerjee:ACL05}, ROUGE-L \cite{lin2004rouge}, CIDEr-D \cite{vedantam2015cider} and SPICE \cite{spice2016}, are leveraged for evaluation.

\textbf{Compared Methods.}
(1) \textbf{LSTM} \cite{Vinyals14} only feeds image into LSTM-based decoder at the initial time step for triggering sentence generation. The reported results are directly drawn from \cite{rennie2017self}. (2) \textbf{SCST} \cite{rennie2017self} devises a self-critical sequence training strategy to train a modified attention-based captioning model in \cite{Xu:ICML15}. (3) \textbf{ADP-ATT} \cite{Xiong2016MetaMind} designs an adaptive attention mechanism to decide whether to attend to the image and which image regions to focus, for image captioning. (4) \textbf{LSTM-A} \cite{yao2017boosting} extends the common encoder-decoder captioning model by additionally injecting semantic attributes into LSTM-based decoder. (5) \textbf{RFNet} \cite{jiang2018recurrent} devises a recurrent fusion network to fuse multiple encoders and generate new informative features for decoder with attention. (6) \textbf{Up-Down} \cite{anderson2017bottom} devises a bottom-up attention mechanism to calculate attention at object level to boost image captioning. (7) \textbf{GCN-LSTM} \cite{yao2018exploring} extends \cite{anderson2017bottom} by exploiting visual relationships between objects. (8) \textbf{Up-Down+HIP} and \textbf{GCN-LSTM+HIP} are our proposals by plugging the devised HIP into Up-Down and GCN-LSTM, respectively. Please note that all the state-of-the-art methods and our models utilize ResNet-101 as the backbone of image encoder, for fair comparison. Besides, we report the results of each model trained with both cross-entropy loss or CIDEr-D reward in self-critical strategy.

\begin{table*}[!tb]\small
  \centering
  \caption{\small Performance (\%) of the top ranking published state-of-the-art image captioning models on the online COCO test server.}
  \label{table:leaderboard}
  \vspace{-0.00in}
  \begin{tabular}{l|*{13}{c|}c}
  \Xhline{2\arrayrulewidth}
      \multicolumn{1}{c|}{\multirow{2}{*}{{Model}}} & \multicolumn{2}{c|}{{BLEU@$1$}} & \multicolumn{2}{c|}{{BLEU@$2$}} & \multicolumn{2}{c|}{{BLEU@$3$}} & \multicolumn{2}{c|}{{BLEU@$4$}} & \multicolumn{2}{c|}{{METEOR}} & \multicolumn{2}{c|}{{ROUGE-L}} & \multicolumn{2}{c}{{CIDEr-D}} \\\cline{2-15}
      \multicolumn{1}{c|}{}&c5 &c40 &c5 &c40 &c5 &c40&c5 &c40&c5 &c40&c5 &c40&c5 &c40 \\\hline
      {GCN-LSTM+HIP}                                 & \textbf{81.6} & \textbf{95.9} & \textbf{66.2} & \textbf{90.4} & \textbf{51.5} & \textbf{81.6} & \textbf{39.3} & \textbf{71.0} & \textbf{28.8} & \textbf{38.1} & \textbf{59.0} & \textbf{74.1} & \textbf{127.9} & \textbf{130.2} \\\hline
      {GCN-LSTM}       \cite{yao2018exploring}                 & 80.8 & 95.2 & 65.5 & 89.3 & 50.8 & 80.3 & 38.7 & 69.7 & 28.5 & 37.6 & 58.5 & 73.4 & 125.3 & 126.5 \\\hline
      {RFNet}       \cite{jiang2018recurrent}               & 80.4 & 95.0 & 64.9 & 89.3 & 50.1 & 80.1 & 38.0 & 69.2 & 28.2 & 37.2 & 58.2 & 73.1 & 122.9 & 125.1 \\\hline
      {Up-Down} \cite{anderson2017bottom} & 80.2 & 95.2 & 64.1 & 88.8 & 49.1 & 79.4 & 36.9 & 68.5 & 27.6 & 36.7 & 57.1 & 72.4 & 117.9 & 120.5  \\\hline
      {LSTM-A} \cite{yao2017boosting}     & 78.7 & 93.7 & 62.7 & 86.7 & 47.6 & 76.5 & 35.6 & 65.2 & 27.0 & 35.4 & 56.4 & 70.5 & 116.0 & 118.0  \\\hline
      {SCST} \cite{rennie2017self}        & 78.1 & 93.7 & 61.9 & 86.0 & 47.0 & 75.9 & 35.2 & 64.5 & 27.0 & 35.5 & 56.3 & 70.7 & 114.7 & 116.7  \\\hline
      \Xhline{2\arrayrulewidth}
  \end{tabular}
  \vspace{-0.18in}
\end{table*}

\subsection{Performance Comparison and Analysis}
\textbf{Performance on COCO.} The performances of different models on COCO for image captioning task are summarized in Table \ref{tab:COCO}. Overall, the results across all metrics and two optimization methods (Cross-Entropy Loss and CIDEr-D Score Optimization) consistently indicate that our GCN-LSTM+HIP exhibits better performances than other approaches, including non-attention models (LSTM, LSTM-A) and attention-based approaches (SCST, ADP-ATT, RFNet, Up-Down, and GCN-LSTM). Up-Down+HIP and GCN-LSTM+HIP by integrating hierarchy parsing architecture makes the absolute improvement over Up-Down and GCN-LSTM by 3.1\% and 3.2\% in terms of CIDEr-D respectively, optimized with cross-entropy loss. The results generally highlight the key advantage of exploiting the hierarchal structure in an image from instance level, region level, to the whole image, pursuing a thorough image understanding for captioning. Specifically, by injecting the high-level semantic attributes into LSTM-based decoder, LSTM-A outperforms LSTM that trains decoder only depending on the input image. Nevertheless, the attention-based methods (SCST, ADP-ATT, Up-Down, and RFNet) exhibit better performance than LSTM-A, which verifies the merit of attention mechanism that dynamically focuses on image regions for sentence generation. Furthermore, GCN-LSTM by exploring the relations between objects to enrich region-level features, improves SCST, ADP-ATT, Up-Down, and RFNet. However, the performances of GCN-LSTM are lower than GCN-LSTM+HIP that additionally exploits hierarchical structure in an image for enhancing all the instance-level, region-level and image-level features and eventually boosting image captioning. In addition, by optimizing the captioning models with CIDEr-D score instead of cross-entropy loss, the CIDEr-D score of GCN-LSTM+HIP is further boosted up to 130.6\%. This confirms that self-critical training strategy is an effective way to amend the discrepancy between training and inference, and improve sentence generation regardless of image captioning approaches. Similar to the observations on the optimization with cross-entropy loss, Up-Down+HIP and GCN-LSTM+HIP lead to better performances than Up-Down and GCN-LSTM when optimized with CIDEr-D score.

\textbf{Ablation Study.} Next, we examine how captioning performance is affected when capitalizing on different features. Table \ref{table:ablation} details the performances by exploiting different features in Up-Down sentence decoder. The use of original region-level features in general achieves a good performance. As expected, only utilizing original instance-level features is inferior to region-level features. The result indicates that the context in the background of a region is still a complement to the foreground. On a throughout hierarchy parsing of an image, the finer features produced by Tree-LSTM lead to better performance. The concatenation of every two features constantly outperforms the individual one. The integration of all three features, i.e., our HIP, reaches the highest performance for captioning. The results basically demonstrate the complementarity in between.

\begin{table}[!tb]\small
  \centering
  \setlength\tabcolsep{1.6pt}
  \caption{\small An ablation study on the use of different features.}
  \label{table:ablation}
  \begin{tabular}{ccc|cccc}
  \Xhline{2\arrayrulewidth}
        Regions      &     Instances    &    Tree-LSTM    &    BLEU@$4$    &     METEOR     &      CIDEr-D    \\ \hline\hline
      $\checkmark$    &                 &                  &    36.2   &   27.0    &    113.5       \\
                     &  $\checkmark$   &                  &    36.1   &   27.0    &    113.3      \\
                     &                 &   $\checkmark$   &    36.3   &   27.4    &    113.7      \\ \hline
     $\checkmark$    &  $\checkmark$   &                  &    36.6   &   27.5    &    114.9     \\
     $\checkmark$    &                 &   $\checkmark$   &    36.8   &   27.9    &    115.5     \\
                     &  $\checkmark$   &   $\checkmark$   &    36.7   &   27.9    &    115.2     \\ \hline
     $\checkmark$    &  $\checkmark$   &   $\checkmark$   &    \textbf{37.0}   &   \textbf{28.1}    &    \textbf{116.6}   \\
  \Xhline{2\arrayrulewidth}
  \end{tabular}
  \vspace{-0.22in}
\end{table}

\begin{figure*}[!tb]
\vspace{-0.02in}
\centering {\includegraphics[width=0.9\textwidth]{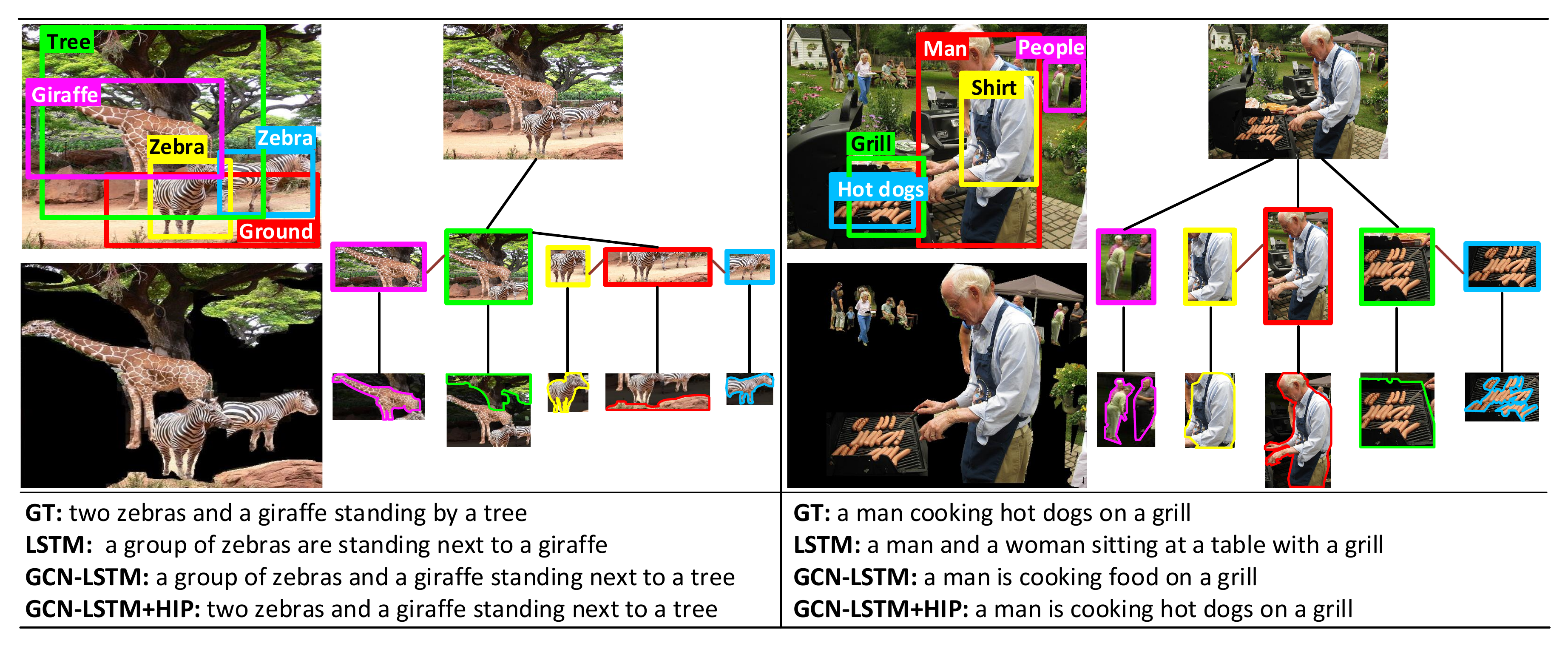}}
\vspace{-0.08in}
\caption{\small Two image examples from COCO dataset with instances, regions, hierarchy structure, and sentence generation results. The output sentences are generated by 1) Ground Truth (GT): One ground truth sentence, 2) LSTM, 3) GCN-LSTM and 4) GCN-LSTM+HIP.}
\vspace{-0.15in}
\label{fig:figRS}
\end{figure*}

\textbf{Qualitative Analysis.} Figure \ref{fig:figRS} showcases a few image examples with instances, regions, the hierarchy structure, ground truth sentences and captions produced by LSTM, GCN-LSTM and GCN-LSTM+HIP, respectively. As illustrated in the exemplar results, the sentences output by GCN-LSTM+HIP are more descriptive. For example, compared to the phrase ``a group of zebras" in the captions produced by LSTM and GCN-LSTM for the first image, the words of ``two zebras" depict the image content more accurately in GCN-LSTM+HIP. We speculate that the results are benefited from the segmentation of two ``zebra" instances and the integration of such information into hierarchy structure. The results again indicate the advantage of guiding sentence generation through holistically interpreting and parsing the structure of an image in our HIP.

\textbf{Performance on COCO Online Testing Server.} We also submitted the run of GCN-LSTM+HIP optimized with CIDEr-D score to online COCO testing server. Table \ref{table:leaderboard} shows the performance Leaderboard on official testing image set with 5 reference captions (c5) and 40 reference captions (c40). Note that here we utilize SENet-154 \cite{hu2018squeeze} as the backbone of Faster R-CNN and Mask R-CNN in our final submission. The latest top-5 performing systems which have been officially published are included in the table. Our GCN-LSTM+HIP leads to performance boost against all the other top-performing systems on the Leaderboard.

\textbf{Human Evaluation.} As the automatic sentence evaluation metrics do not necessarily correlate with human judgement, we additionally conducted a human study to evaluate GCN-LSTM+HIP against two baselines, i.e., LSTM and GCN-LSTM. We invite 12 labelers and randomly select 1K images from testing set for human evaluation. All the labelers are grouped into two teams. We show the first team each image with three auto-generated sentences plus three human-annotated captions and ask the labelers: Do the systems produce human-like sentences? Instead, we show the second team only one sentence at a time, which could be generated by captioning methods or human annotation (Human). The labelers are asked: Can you distinguish human annotation from that by a system? Based on labelers' feedback, we calculate two metrics: 1) M1: percentage of captions that are as well as or even better than human annotation; 2) M2: percentage of captions that pass the Turing Test. The M1 scores of GCN-LSTM+HIP, GCN-LSTM and LSTM are 76.5\%, 73.9\% and 50.7\%, respectively. In terms of M2, Human, GCN-LSTM+HIP, GCN-LSTM, and LSTM achieve 91.4\%, 85.2\%, 81.5\%, and 57.1\%. Apparently, our GCN-LSTM+HIP is the winner on both criteria.

\begin{figure}[!tb]
\vspace{-0.05in}
\centering {\includegraphics[width=0.4\textwidth]{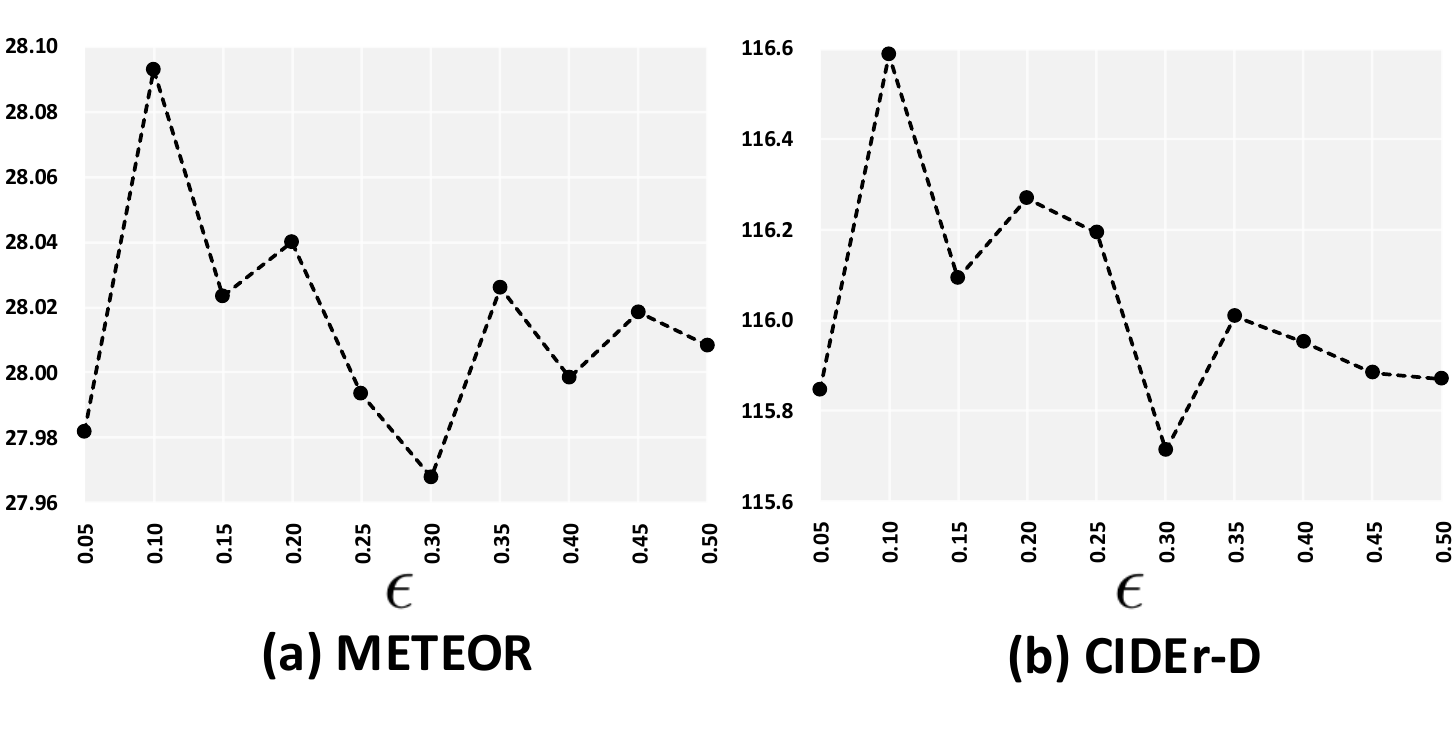}}
\vspace{-0.15in}
\caption{The effect of the threshold parameter $\epsilon$ for constructing hierarchy in Up-Down+HIP with cross-entropy loss over (a) METEOR (\%) and (b) CIDEr-D (\%) on COCO.}
\label{fig:figtradeoff}
\vspace{-0.25in}
\end{figure}

\textbf{Effect of the threshold $\epsilon$.} To clarify the effect of the threshold parameter $\epsilon$ for constructing hierarchy, we illustrate the performance curves over METEOR and CIDEr-D with different threshold parameters in Figure \ref{fig:figtradeoff}. As shown in the figure, we can see that both performance curves are generally like the ``$\wedge$" shapes when $\epsilon$ varies in a range from 0.05 to 0.5. Hence we set the threshold parameter $\epsilon$ as 0.1 in our experiments, which can achieves the best performance.

\textbf{Extension to Recognition.} As a feature refiner, here we test the generalizability of our HIP on recognition task. We also experiment with COCO dataset on 80 object categories and utilize object annotations as multiple labels of an image. Multi-label softmax loss \cite{guillaumin2009tagprop} is exploited for classification. For each image, we predict top-3 ranked labels. Then, we compute the precision and recall for each label separately, and report per-class precision (C-P) and pre-class recall (C-R). Furthermore, to alleviate the bias towards infrequent labels, we also compute overall precision (O-P) and overall recall (O-R). As the harmonic average of precision and recall, F1 (C-F1 and O-F1) scores are given as well. Table \ref{table:recognition} details the performances of different features on recognition task. We take image-level features from our HIP and compare to mean-pooled image features in Up-Down \cite{anderson2017bottom}. By delving into hierarchy parsing, image-level features from HIP lead to 2\% and 3.2\% performance gain in C-F1 and O-F1 over the features in Up-Down. The results basically verify the generalizability of HIP on recognition task.

\begin{table}[!tb]\small
\vspace{-0.02in}
  \centering
  \caption{\small Performance comparisons on recognition task when employing different features.}
  \label{table:recognition}
  \begin{tabular}{c|c@{~}c@{~}c|c@{~}c@{~}c}
  \Xhline{2\arrayrulewidth}
         &   C-P  ~  &   C-R   ~ &   C-F1    &    O-P  ~  &    O-R  ~  &    O-F1 \\ \hline
     Up-Down \cite{anderson2017bottom} &   65.32 ~  &   62.24 ~  &   63.74    &   64.37  ~ &   66.48 ~  &   65.41 \\
     HIP       &   \textbf{66.18} ~  &   \textbf{65.30}  ~ &   \textbf{65.74}    &   \textbf{67.53}  ~ &   \textbf{69.74} ~ &  \textbf{68.61} \\
  \Xhline{2\arrayrulewidth}
  \end{tabular}
  \vspace{-0.25in}
\end{table}

\section{Conclusions}
We have presented HIerarchy Parsing (HIP) architecture, which integrates hierarchical structure into image encoder to boost captioning. Particularly, we study the problem from the viewpoint of interpreting the hierarchy in tree-structured topologies from the root of the whole image to the middle layers of regions and finally to the leaf layer of instances. To verify our claim, we have built a three-level hierarchal structure of constituent visual patterns (i.e., instances, regions and the whole image) in an image. A Tree-LSTM is employed on the hierarchy to enrich the features throughout all the three levels. Extensive experiments conducted on COCO image captioning dataset demonstrate the efficacy of HIP in both cases of feeding a blend of features directly from HIP or further enhanced version with relation into an attention-based sentence decoder. More remarkably, we achieve new state-of-the-art performances on this captioning dataset. The evaluations on HIP also validate its potential of generalizing to recognition task.

{\small
\bibliographystyle{ieee_fullname}
\bibliography{egbib}
}

\end{document}